\documentclass[sigconf]{acmart}

\AtBeginDocument{%
  }

\usepackage{bm}
\usepackage{multirow}
\usepackage{graphicx}
\usepackage{subcaption}
\usepackage{verbatim}
\usepackage{adjustbox}
\usepackage{algorithm}
\usepackage{algorithmic}

\copyrightyear{2026}
\acmYear{2026}
\setcopyright{cc}
\setcctype{by}
\acmConference[GECCO Companion '26]{Genetic and Evolutionary Computation Conference}{July 13--17, 2026}{San Jose, Costa Rica}
\acmBooktitle{Genetic and Evolutionary Computation Conference (GECCO Companion '26), July 13--17, 2026, San Jose, Costa Rica}
\acmDOI{10.1145/3795101.3805285}
\acmISBN{979-8-4007-2488-6/2026/07}

\begin{document}
\title[GeM-EA: A Generative and Meta-learning Enhanced EA]{GeM-EA: A Generative and Meta-learning Enhanced Evolutionary Algorithm for Streaming Data-Driven Optimization}

\author{Yue Wu$^{*}$, Yuan-Ting Zhong$^{*}$, Ze-Yuan Ma, Yue-Jiao Gong$^{\dagger}$}

\affiliation{
  \institution{South China University of Technology}
  \city{Guangzhou}
  \country{China}
}

\affiliation{
  \institution{\vspace{0.1cm} \normalsize $^{*}$Equal contribution. $^{\dagger}$Corresponding author: gongyuejiao@gmail.com}
  \country{} 
}

\begin{abstract}
Streaming Data-Driven Optimization (SDDO) problems arise in many applications where data arrive continuously and the optimization environment evolves over time. Concept drift produces non-stationary landscapes, making optimization methods challenging due to outdated models. Existing approaches often rely on simple surrogate combinations or directly injecting solutions, which may cause negative transfer under sudden environmental changes. We propose GeM-EA, a Generative and Meta-learning Enhanced Evolutionary Algorithm for SDDO that unifies meta-learned surrogate adaptation with generative replay for effective evolutionary search. Upon detecting concept drift, a bi-level meta-learning strategy rapidly initializes the surrogate using environment-relevant priors, while a linear residual component captures global trends. A multi-island evolutionary strategy further leverages historical knowledge via generative replay to accelerate optimization. Experimental results on benchmark SDDO problems demonstrate that GeM-EA achieves faster adaptation and improved robustness compared with state-of-the-art methods.The source code is available at https://github.com/PoetMoon/GeM-EA.
\end{abstract}

\ccsdesc[500]{Theory of computation~Evolutionary algorithms}
\ccsdesc[500]{Information systems~Data streams}

\keywords {Streaming data-driven evolutionary algorithms, Data stream, Meta-learning, Concept drift}

\maketitle

\section{Introduction}

With ongoing technological advances, applications like smart city management~\cite{chen2021trafficstream,peixoto2023fogjam} and intelligent transportation systems~\cite{osekowska2017maritime} increasingly rely on continuous data streams. Optimization in such settings is inherently dynamic, as the underlying data distribution evolves over time—a phenomenon known as concept drift~\cite{widmer1996learning}. This context defines Streaming Data-Driven Optimization (SDDO) problems~\cite{zhong2024sddobench}, requiring methods to adapt dynamically to maintain performance.

To address SDDO, Streaming Data-Driven Evolutionary Algorithms (SDDEAs) integrate Evolutionary Algorithms (EAs) with data-driven modeling~\cite{gong2023offline}. While existing methods, such as explicit historical solution retrieval~\cite{zhong2025data}, attempt to exploit past knowledge, current approaches still suffer from rigid knowledge reuse. Specifically, linear surrogate combinations fail to capture complex geometric transformations like rotation. Such spatial shifts introduce severe asymmetry and non-separability into the optimization landscape. Consequently, injecting individuals based on these misaligned models creates a high risk of negative transfer, misleading the search under sudden drifts.

Meta-learning~\cite{finn2017model} offers a promising ``learning to adapt'' paradigm that transcends linear ensembles by learning malleable initializations for complex landscape changes. Despite success in general machine learning~\cite{ravi2017optimization,vinyals2016matching}, meta-learning remains largely unexplored in SDDO~\cite{zhang2024solving}. Existing attempts require active data sampling, a practice strictly prohibited in SDDO. Furthermore, traditional meta-learning assumes tasks are sampled from a stationary distribution~\cite{finn2019online} and relies on discrete task formulations. These assumptions fundamentally clash with SDDO's continuous streaming nature, where concept drift is unpredictable and explicit environment segmentation is unavailable.

To overcome these limitations, we propose \textbf{GeM-EA}, a \emph{Generative and Meta-learning Enhanced Evolutionary Algorithm} for SDDO. Our main contributions are:
\begin{figure*}[t]
    \centering
    \includegraphics[trim=0.0cm 0.0cm 0.0cm 0.0cm, clip, width=0.8\linewidth]{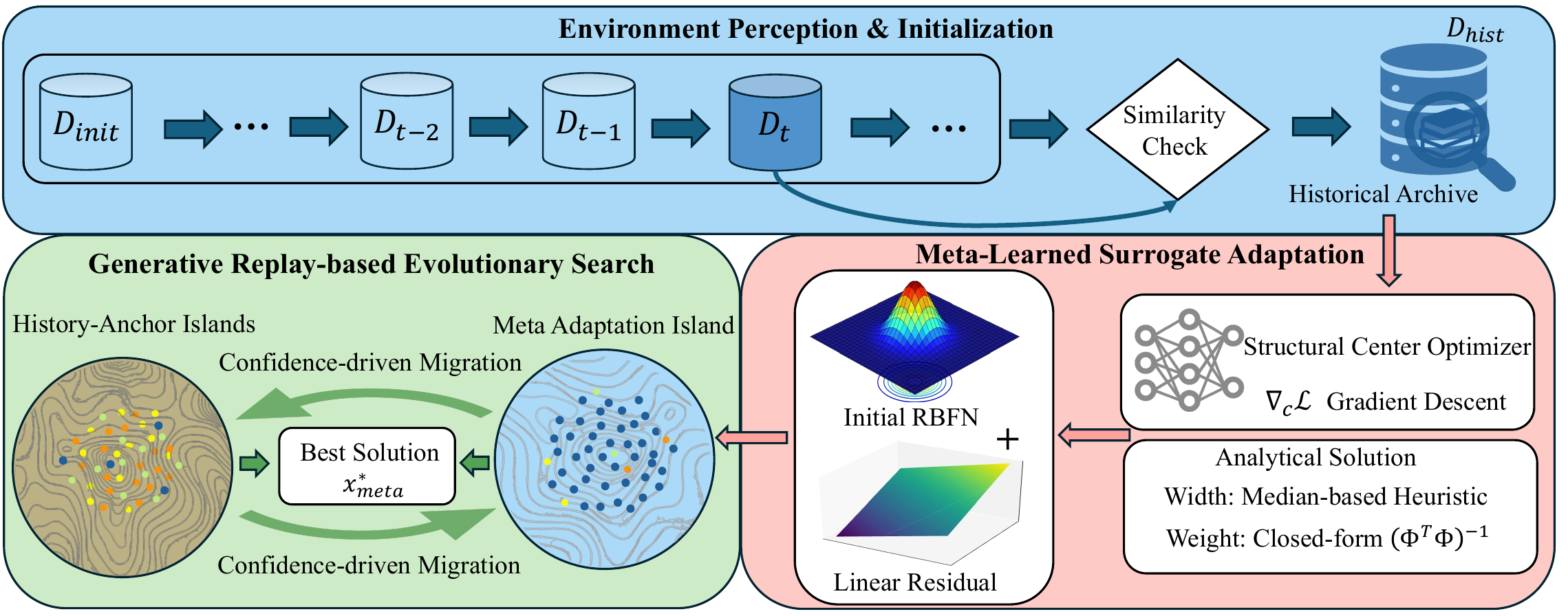}
    \caption{The framework of GeM-EA}
    \label{fig:framework}
   
\end{figure*}
\begin{itemize}

\item A unified framework synergizing \emph{meta-learned surrogate adaptation} with \emph{generative replay-based evolutionary search}. It rapidly initializes the surrogate upon concept drift \cite{zhong2026trace} without active data acquisition and distills historical solutions via generative replay, accelerating search and mitigating negative transfer.

\item A bi-level \emph{meta-learned surrogate adaptation} module. It decouples structural topology adaptation from fast adaptation by meta-optimizing structural parameters while analytically solving non-structural widths and weights. This alleviates instability in fully gradient-based meta-learning, while a linear residual component further captures global landscape trends.

\item A \emph{generative replay-based evolutionary search} utilizing a multi-island architecture. A meta-adaptation island aggressively explores the current landscape, while history-anchor islands maintain robustness using generated historical solutions. A confidence-driven migration mechanism coordinates their exchange, isolating the search from misleading historical biases.

\item Extensive experiments demonstrate that \textbf{GeM-EA} consistently achieves superior solution quality, faster adaptation, and high computational efficiency compared with state-of-the-art SDDO methods.

\end{itemize}

\section{Methodology}
As illustrated in Figure \ref{fig:framework}, GeM-EA synergizes meta-learned surrogate adaptation with generative replay-based evolutionary search. We employ a Radial Basis Function Network (RBFN)~\cite{park1991universal} as the base surrogate, uniquely characterized by parameter separability into structural parameters ($\theta$) and non-structural parameters ($w, \sigma$). 

\subsection{Meta-Learned Surrogate Adaptation}
Directly optimizing the parameter set $\Psi=\{\theta,w,\sigma\}$ in a unified meta-learning loop is computationally unstable due to severe gradient conflicts. This stems from: 1) \textit{convexity mismatch}, as weights $w$ follow a convex least-squares objective while structural parameters $\theta$ occupy a highly non-convex landscape; and 2) \textit{sensitivity disparity}, since minor $\theta$ perturbations fundamentally reshape basis functions, whereas $w$ changes only induce linear scaling. Consequently, joint updates frequently trigger pronounced oscillations or premature stagnation.

To overcome this, we propose a bi-level meta-learned surrogate adaptation strategy that decouples structural parameter optimization from fast adaptation, enabling stable surrogate initialization and reliable meta-adaptation under streaming environments.

\textit{Stage 1: Environment-Relevant Prior Identification.} To mitigate negative transfer in streaming environments, we construct a meta-dataset $\mathcal{S}_{meta}$ by retrieving only historical environments highly relevant to the current dataset $D_t$. Relevance is evaluated via a discrepancy metric combining approximation error and prediction divergence: 
\begin{equation}
    Dist(D_t, \mathcal{M}_i) = \gamma_1 \text{MAPE}(D_t, \mathcal{M}_i) + \gamma_2 \frac{1}{N} \sum_{\mathbf{x} \in D_t} || \mathcal{M}_{temp}(\mathbf{x}) - \mathcal{M}_i(\mathbf{x}) ||^2
\end{equation}
where $\text{MAPE}$ denotes mean absolute percentage error, and $\mathcal{M}_{temp}$ is a temporary surrogate trained exclusively on $D_t$ serving as a structural baseline.

\textit{Stage 2: Bi-Level Meta-Learned Fast Adaptation.} Environment-level structural basis parameters $\theta$ update via gradient descent for topological shifts ($\theta^{\prime}_t = \theta - \alpha \nabla_{\theta} \mathcal{L}_{D_t}$), and widths $\sigma^{\prime}_t$ recalibrate via $k$-nearest neighbor heuristics. Output weights $w^{\prime}_t$ are analytically resolved via Ridge Regression using the updated basis activation matrix $\Phi$ and ground-truth $y_t$:
\begin{equation}
    w^{\prime}_t = \left( \Phi(\theta^{\prime}_t, \sigma^{\prime}_t)^T \Phi(\theta^{\prime}_t, \sigma^{\prime}_t) + \lambda \mathbf{I} \right)^{-1} \Phi(\theta^{\prime}_t, \sigma^{\prime}_t)^T y_t
\end{equation}
Meta-level $\theta$ is gradient-refined minimizing cumulative generalization loss across $\mathcal{S}_{meta}$ ($\theta \leftarrow \theta - \beta \nabla_{\theta} \sum_{D_t} \mathcal{L}_{D_t}$), decoupling structural tuning from analytical weight solving.

\textit{Stage 3: Residual-Augmented Stabilization.} To mitigate limited-data residual approximation errors, the final surrogate augments the meta-learned $\hat{f}_{meta}(\mathbf{x})$ with a global linear residual: $\hat{F}(\mathbf{x}) = \hat{f}_{meta}(\mathbf{x}) + \mathbf{a}^T \mathbf{x} + b$. Linear parameters $\{\mathbf{a}, b\}$ are analytically solved via pseudo-inverse to minimize the residual Mean Squared Error ($e_j = y_j - \hat{f}_{meta}(\mathbf{x}_j)$), efficiently balancing fast environment-specific adaptation and global stability.

\subsection{Generative Replay-based Evolutionary Search}

Instead of relying on a single population vulnerable to negative transfer, GeM-EA employs a multi-island architecture. The \textit{meta-adaptation island} conducts aggressive exploration uniformly guided by the meta-learned surrogate. Simultaneously, $P$ \textit{history-anchor islands} are generated from the statistical summaries of the top-$P$ most similar past environments, acting as stabilizing anchors to accelerate exploitation. 

\begin{algorithm}[htbp]
\caption{Generative Replay-based Evolutionary Search}
\label{alg:GeM_ES}
\small
\linespread{0.9}\selectfont %
\begin{algorithmic}[1]
\renewcommand{\algorithmicrequire}{\textbf{Input:}}
\renewcommand{\algorithmicensure}{\textbf{Output:}}

\REQUIRE History Archive $Arc$, Migration interval $\tau$, Max generations $G_{max}$, Meta-learned surrogate $\hat{F}(\mathbf{x})$
\ENSURE Optimal solution $\mathbf{x}^*_{meta}$

\STATE \textbf{Initialization:}
\STATE Select top-$P$ $\mathcal{S}_{meta}$ from $Arc$
\STATE Randomly initialize meta-island $\Lambda_{meta}$ 
\STATE Generate anchor islands $\{\Lambda_{anchor}^{(i)}\}_{i=1}^P$ based on $\mathcal{S}_{meta}$

\STATE \textbf{Evolutionary Search:}
\FOR{$g = 1$ to $G_{max}$}
    \STATE \textit{// Meta-Adaptation Island}
    \STATE $\Lambda_{meta} \leftarrow \text{Evolve}(\Lambda_{meta}, \hat{F}_{meta})$
    \STATE $\mathbf{x}^*_{meta} \leftarrow \text{Best}(\Lambda_{meta})$

    \STATE \textit{// History-Anchor Islands}
    \FOR{$i = 1$ to $P$}
        \STATE $\Lambda_{anchor}^{(i)} \leftarrow \text{Evolve}(\Lambda_{anchor}^{(i)}, \hat{F}_{i})$
        \STATE $\mathbf{x}^*_{i} \leftarrow \text{Best}(\Lambda_{anchor}^{(i)})$
    \ENDFOR

    \STATE \textit{// Bidirectional Migration}
    \IF{$g \mod \tau = 0$}
        \STATE $\Lambda_{meta} \leftarrow \Lambda_{meta} \cup \{\mathbf{x}^*_{1}, \dots, \mathbf{x}^*_{P}\}$ 
        \FOR{$i = 1$ to $P$}
            \IF{$\hat{F}_{i}(\mathbf{x}^*_{meta}) < \hat{F}_{i}(\mathbf{x}^*_{i})$}
                \STATE $\Lambda_{anchor}^{(i)} \leftarrow \Lambda_{anchor}^{(i)} \cup \{\mathbf{x}^*_{meta}\}$
            \ENDIF
        \ENDFOR
    \ENDIF
\ENDFOR

\STATE Update Archive $Arc$ based on clustering
\RETURN $\mathbf{x}^*_{meta}$

\end{algorithmic}
\end{algorithm}

To coordinate these islands, we implement a confidence-driven migration mechanism. At regular intervals $\tau$, elite solutions from history-anchor islands are injected into the meta-adaptation island. Conversely, if the meta-adaptation island discovers superior basins of attraction, its best solution $\mathbf{x}^*_{meta}$ is fed back to an anchor island $i$ only if it satisfies a strict confidence condition: $\hat{F}_{i}(\mathbf{x}^*_{meta}) < \hat{F}_{i}(\mathbf{x}^*_{i})$. This bidirectional exchange balances exploration and historical exploitation while effectively shielding the search from negative transfer.

\section{Experimental Results}

\subsection{Experimental Settings}
We utilize SDDObench~\cite{zhong2024sddobench} . Algorithm performance is evaluated using two standard metrics: offline error ($E_{offline}$) and online error($E_{online}$). Detailed mathematical definitions of these metrics, along with extended related work on SDDO, are provided in the Supplementary Appendix 1. To ensure strictly fair comparisons, all evaluated algorithms (TT-DDEA~\cite{huang2021offline}, BDDEA-LDG~\cite{li2020boosting}, MLO~\cite{zhang2024solving}, DETO~\cite{li2023data}, DSE-MFS~\cite{yang2023data}, SAEF-1GP~\cite{luo2018surrogate}, and DASE~\cite{zhong2025data}) share the same evolutionary budget: a population size of 300 and a maximum of 50 iterations per environment. For GeM-EA, the RBFN uses $K_c=\lfloor\sqrt{N_{data}}\rfloor$ centers, $\lambda=0.01$, and a meta-learning rate of $10^{-4}$. The DE optimizer uses DE/current-to-best/1 ($F=0.5, Cr=0.9$) with $P=4$ history-anchor islands and a migration interval $\tau=10$.
To ensure statistical robustness, each experiment is conducted 10 times independently; performance is reported as the mean $\pm$ standard deviation calculated from these repetitions.

Statistical significance is rigorously assessed using the Kruskal-Wallis test followed by the post hoc Dunnett’s test with a Bonferroni correction strategy. The significance level is 0.05. 
To clearly visualize the comparative performance in the result tables, we use "$+$" to denote that our method significantly outperforms the competitor, "$\approx$" to indicate statistically similar performance, and "$-$" to denote underperformance.
\subsection{Comparative Analysis with State-of-the-Art Methods}
\subsubsection{Comparative Solution Quality Analysis.} As shown in Table \ref{tab:offline_results}, GeM-EA achieves the lowest average rank (1.22) across all metrics, establishing a significant lead. The advantage is pronounced on complex multimodal landscapes (e.g., F5-D3), where GeM-EA reduces offline error by an order of magnitude compared to MLO.

\begin{figure}[h!]
    \centering
    \includegraphics[width=\linewidth]{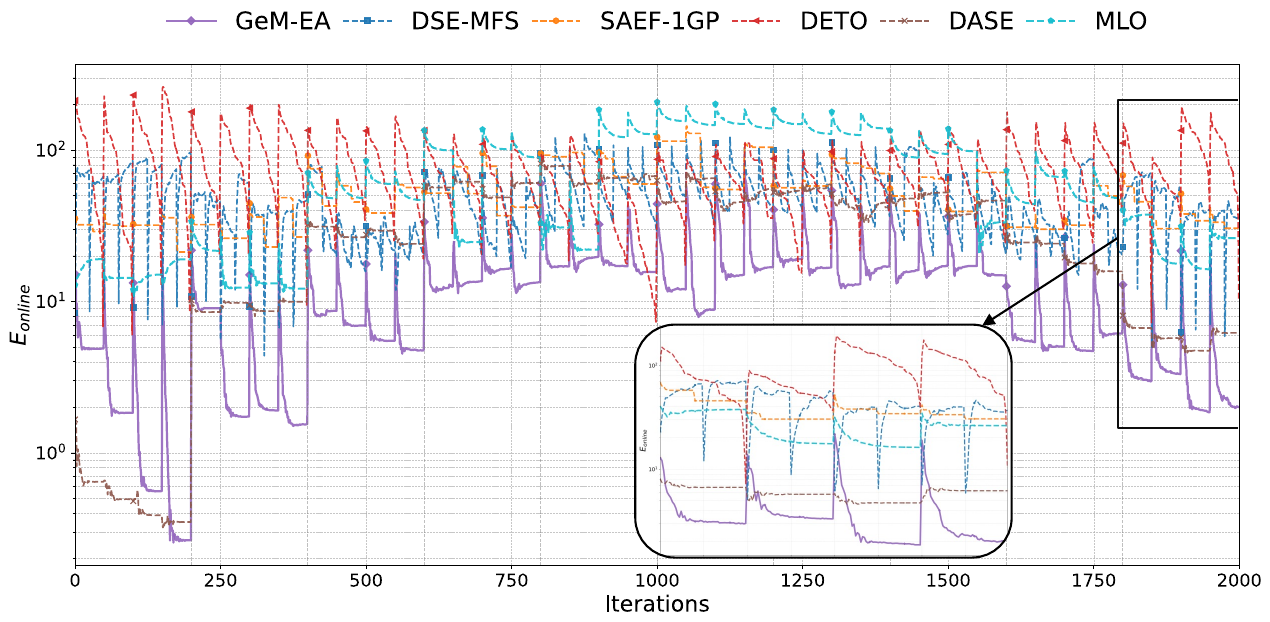}
    \caption{Online Error convergence trajectories on the last ten environments of SDDObenchF4-D4}
    \label{fig:online_error_comparison}
    
\end{figure}
\begin{table*}[htbp]
    \centering
    \caption{The Mean and Standard Deviation of $E_{offline}$ on SDDObench}
    \label{tab:offline_results}
    \resizebox{0.85\linewidth}{!}{
        \begin{tabular}{|c|c||c|c||c|c|c|c|c||c|}
        \hline
        \multirow{2}{*}{\textbf{Instance}} & \multirow{2}{*}{\textbf{Drift}} & \multicolumn{2}{c||}{\textbf{DDEAs}} & \multicolumn{5}{c||}{\textbf{SDDEAs}} & \textbf{Proposed} \\ \cline{3-10}
         & & \textbf{TT-DDEA} & \textbf{BDDEA-LDG} & \textbf{MLO} & \textbf{DETO} & \textbf{DSE-MFS} & \textbf{SAEF-1GP} & \textbf{DASE} & \textbf{GeM-EA} \\
        \hline
        \hline
        
        \multirow{5}{*}{F1} 
         & D1 & 6.82e+01$\pm$2.15e-01 (+)& 6.77e+01$\pm$1.41e-02 (+)& 6.82e+01$\pm$1.50e-01 (+)& 6.79e+01$\pm$2.29e-01 (+)& 6.77e+01$\pm$2.37e-02 (+)& 6.75e+01$\pm$4.94e-03 (+)& 6.76e+01$\pm$1.90e-01 (+)& \textbf{6.60e+01$\pm$2.70e-01} \\ \cline{2-10}
         & D2 & 6.70e+01$\pm$2.03e-01 (+)& 6.78e+01$\pm$6.72e-03 (+)& 6.71e+01$\pm$8.00e-02 (+)& 6.66e+01$\pm$4.08e-02 (+)& 6.53e+01$\pm$1.95e-02 (+)& 6.76e+01$\pm$4.61e-03 (+)& 6.63e+01$\pm$1.20e-01 (+)& \textbf{6.45e+01$\pm$2.30e-01} \\ \cline{2-10}
         & D3 & 6.59e+01$\pm$1.97e-01 (+)& 6.52e+01$\pm$2.05e-02 (+)& 6.60e+01$\pm$1.20e-01 (+)& 6.54e+01$\pm$1.14e-01 (+)& 6.56e+01$\pm$9.93e-03 (+)& 6.75e+01$\pm$6.97e-03 (+)& 6.48e+01$\pm$2.10e-01 (+)& \textbf{6.25e+01$\pm$6.00e-01} \\ \cline{2-10}
         & D4 & 6.51e+01$\pm$1.85e-01 (+)& 6.46e+01$\pm$2.20e-02 (+)& 6.51e+01$\pm$9.00e-02 (+)& 6.45e+01$\pm$9.70e-02 (+)& 6.50e+01$\pm$7.34e-03 (+)& 6.56e+01$\pm$2.33e-03 (+)& 6.46e+01$\pm$1.00e-01 (+)& \textbf{6.31e+01$\pm$1.10e-01} \\ \cline{2-10}
         & D5 & 6.53e+01$\pm$1.92e-01 (+)& 6.47e+01$\pm$2.14e-02 (+)& 6.53e+01$\pm$1.10e-01 (+)& 6.48e+01$\pm$2.47e-01 (+)& 6.51e+01$\pm$2.61e-02 (+)& 6.46e+01$\pm$2.56e-03 (+)& 6.48e+01$\pm$1.00e-01 (+)& \textbf{6.29e+01$\pm$1.30e-01} \\
        \hline 
        
        \multirow{5}{*}{F2} 
         & D1 & 6.82e+01$\pm$2.11e-01 (+)& 6.77e+01$\pm$4.28e-03 ($\approx$)& 6.82e+01$\pm$1.80e-01 (+)& 6.79e+01$\pm$2.41e-01 (+)& 6.77e+01$\pm$6.33e-02 ($\approx$)& 6.75e+01$\pm$ 1.84e-03 ($\approx$)& 6.76e+01$\pm$1.50e-01 ($\approx$)& \textbf{6.71e+01$\pm$3.20e-01} \\ \cline{2-10}
         & D2 & 6.68e+01$\pm$1.94e-01 (+)& 6.76e+01$\pm$6.27e-01 (+)& 6.68e+01$\pm$2.50e-01 (+)& 6.76e+01$\pm$3.05e-01 (+)& 6.74e+01$\pm$3.70e-01 (+)& 6.76e+01$\pm$4.09e-03 (+)& 6.71e+01$\pm$2.10e-01 (+)& \textbf{6.51e+01$\pm$3.80e-01} \\ \cline{2-10}
         & D3 & 6.66e+01$\pm$1.88e-01 (+)& 6.65e+01$\pm$4.85e-01 (+)& 6.71e+01$\pm$3.20e-01 (+)& 6.67e+01$\pm$5.95e-02 (+)& 6.63e+01$\pm$3.12e-01 (+)& 6.75e+01$\pm$5.46e-03 (+)& 6.61e+01$\pm$7.90e-01 (+)& \textbf{6.41e+01$\pm$5.90e-01} \\ \cline{2-10}
         & D4 & 6.64e+01$\pm$1.96e-01 ($\approx$)& 6.66e+01$\pm$3.44e-01 ($\approx$)& 6.79e+01$\pm$4.10e-01 (+)& 6.63e+01$\pm$1.59e-01 ($\approx$)& 6.66e+01$\pm$1.04e+00 ($\approx$)& 6.76e+01$\pm$ 3.53e-03 (+)& 6.66e+01$\pm$7.20e-01 ($\approx$)& \textbf{6.61e+01$\pm$7.20e-01} \\ \cline{2-10}
         & D5 & 6.77e+01$\pm$2.08e-01 (+)& 6.63e+01$\pm$2.26e-01 ($\approx$)& 6.63e+01$\pm$2.80e-01 ($\approx$)& 6.61e+01$\pm$4.19e-02 ($\approx$)& 6.69e+01$\pm$9.16e-01 (+)& 6.75e+01$\pm$ 4.63e-03 (+)& 6.70e+01$\pm$8.40e-01 (+)& \textbf{6.58e+01$\pm$7.90e-01} \\
        \hline 
        
        \multirow{5}{*}{F3} 
         & D1 & 6.82e+01$\pm$2.15e-01 (+)& 6.77e+01$\pm$1.82e-02 (+)& 6.82e+01$\pm$1.60e-01 (+)& 6.79e+01$\pm$1.01e-01 (+)& 6.78e+01$\pm$1.83e-02 (+)& 6.75e+01$\pm$5.27e-03 (+)& 6.76e+01$\pm$2.10e-01 (+)& \textbf{6.62e+01$\pm$2.10e-01} \\ \cline{2-10}
         & D2 & 6.66e+01$\pm$1.90e-01 (+)& 6.73e+01$\pm$1.33e-02 (+)& 6.66e+01$\pm$2.10e-01 (+)& 6.61e+01$\pm$1.83e-01 (+)& \textbf{6.37e+01 $\pm$1.10e-02} (-)& 6.76e+01$\pm$3.03e-03 (+)& 6.60e+01$\pm$1.50e-01 ($\approx$)& 6.54e+01$\pm$3.70e-01 \\ \cline{2-10}
         & D3 & 6.35e+01$\pm$1.72e-01 (+)& 6.33e+01$\pm$1.03e-01 (+)& 6.35e+01$\pm$1.50e-01 (+)& 6.29e+01$\pm$3.90e-02 (+)& 6.35e+01$\pm$6.56e-02 (+)& 6.35e+01 $\pm$ 3.99e-03 (+)& 6.28e+01$\pm$9.00e-02 (+)& \textbf{6.18e+01$\pm$2.70e-01} \\ \cline{2-10}
         & D4 & 6.33e+01$\pm$1.65e-01 (+)& 6.31e+01$\pm$1.16e-01 (+)& 6.32e+01$\pm$1.80e-01 (+)& 6.25e+01$\pm$8.72e-02 (+)& 6.34e+01 $\pm$ 3.74e-02 (+)& 6.35e+01 $\pm$ 3.53e-03 (+)& 6.28e+01$\pm$1.20e-01 (+)& \textbf{6.14e+01$\pm$1.90e-01} \\ \cline{2-10}
         & D5 & 6.35e+01$\pm$1.68e-01 (+)& 6.30e+01$\pm$9.22e-02 (+)& 6.33e+01$\pm$1.40e-01 (+)& 6.28e+01$\pm$2.15e-01 (+)& 6.34e+01$\pm$1.41e-01 (+)& 6.35e+01$\pm$ 6.15e-03 (+)& 6.28e+01$\pm$1.80e-01 (+)& \textbf{6.12e+01$\pm$1.60e-01} \\
        \hline 

        \multirow{5}{*}{F4} 
         & D1 & 1.86e+02$\pm$4.85e+00 (+)& 4.21e+00$\pm$3.41e-02 (+)& 8.79e+01$\pm$2.85e+00 (+)& 1.08e+02$\pm$3.34e+00 (+)& 6.73e+01$\pm$2.88e-01 (+)& 5.04e+01$\pm$2.80e-01 (+)& 1.50e-01$\pm$0.03e-01 (+)& \textbf{6.00e-02$\pm$1.00e-02} \\ \cline{2-10}
         & D2 & 1.50e+02$\pm$3.76e+00 (+)& 2.15e+01$\pm$4.06e-01 (+)& 1.01e+02$\pm$3.40e+00 (+)& 9.87e+01$\pm$2.41e+00 (+)& 4.35e+01$\pm$3.09e-01 (+)& 6.22e+01$\pm$ 1.30e+00 (+)& 1.46e+01$\pm$3.38e+00 (+)& \textbf{6.00e+00$\pm$9.60e-01} \\ \cline{2-10}
         & D3 & 1.21e+02$\pm$3.15e+00 (+)& 2.20e+01$\pm$2.47e-01 (+)& 1.05e+02$\pm$4.15e+00 (+)& 8.62e+01$\pm$2.74e+00 (+)& 4.28e+01 $\pm$3.71e+00 (+)& 6.08e+01$\pm$ 2.90e-01 (+)& 1.71e+01$\pm$3.73e+00 (+)& \textbf{5.35e+00$\pm$5.80e-01} \\ \cline{2-10}
         & D4 & 1.13e+02$\pm$2.84e+00 (+)& 4.21e+01$\pm$4.92e-01 (+)& 1.19e+02$\pm$5.20e+00 (+)& 7.82e+01$\pm$2.83e+00 (+)& 5.45e+01$\pm$3.39e+00 (+)& 7.23e+01$\pm$6.43e-01 (+)& 2.99e+01$\pm$5.19e+00 (+)& \textbf{1.01e+01$\pm$1.34e+00} \\ \cline{2-10}
         & D5 & 1.20e+02$\pm$3.02e+00 (+)& 4.01e+01$\pm$1.09e+00 (+)& 1.17e+02$\pm$4.80e+00 (+)& 8.40e+01$\pm$1.69e+00 (+)& 6.24e+01$\pm$2.56e+00 (+)& 7.40e+01$\pm$ 1.67e-01 (+)& 3.77e+01$\pm$4.83e+00 (+)& \textbf{1.37e+01$\pm$2.24e+00} \\
        \hline 

        \multirow{5}{*}{F5} 
         & D1 & 4.20e+03$\pm$1.05e+02 (+)& 1.31e+02$\pm$3.15e+00 (+)& 4.56e+03$\pm$1.45e+02 (+)& 6.34e+03$\pm$1.78e+02 (+)& 2.81e+03$\pm$1.02e+02 (+)& 2.09e+03$\pm$1.20e+01 (+)& \textbf{4.45e+01$\pm$7.56e+00} (-)& 5.62e+01$\pm$6.30e+00 \\ \cline{2-10}
         & D2 & 4.57e+03$\pm$1.14e+02 (+)& 6.86e+02 $\pm$1.65e+01 (+)& 4.84e+03$\pm$1.68e+02 (+)& 5.00e+03$\pm$1.35e+02 (+)& 2.01e+03$\pm$7.83e+01 (+)& 3.68e+03$\pm$ 4.98e+01 (+)& 5.80e+02$\pm$1.50e+02 (+)& \textbf{3.65e+02$\pm$4.16e+01} \\ \cline{2-10}
         & D3 & 5.45e+03$\pm$1.36e+02 (+)& 6.39e+02 $\pm$1.76e+01 (+)& 5.92e+03$\pm$2.10e+02 (+)& 4.73e+03$\pm$1.11e+02 (+)& 2.85e+03$\pm$1.80e+02 (+)& 2.95e+03$\pm$2.51e+01 (+)& 4.60e+02$\pm$4.30e+01 (+)& \textbf{2.59e+02$\pm$3.05e+01} \\ \cline{2-10}
         & D4 & 5.36e+03$\pm$1.34e+02 (+)& 1.72e+03$\pm$5.47e+01 (+)& 7.13e+03$\pm$2.55e+02 (+)& 4.42e+03$\pm$9.52e+01 (+)& 2.99e+03$\pm$9.44e+01 (+)& 3.90e+03$\pm$7.48e+01 (+)& 1.44e+03$\pm$3.41e+02 (+)& \textbf{7.12e+02$\pm$1.46e+02} \\ \cline{2-10}
         & D5 & 5.46e+03$\pm$1.37e+02 (+)& 1.77e+03$\pm$5.76e+01 (+)& 7.07e+03$\pm$2.40e+02 (+)& 4.62e+03$\pm$1.37e+02 (+)& 2.91e+03 $\pm$ 1.34e+02 (+)& 4.04e+03$\pm$1.74e+01 (+)& 1.58e+03$\pm$3.91e+02 (+)& \textbf{8.18e+02$\pm$1.35e+02} \\
        \hline 

        \multirow{5}{*}{F6} 
         & D1 & 2.10e+01$\pm$5.24e-01 (+)& 1.15e+01$\pm$3.27e-02 (+)& 2.11e+01$\pm$4.50e-01 (+)& 2.07e+01$\pm$4.83e-01 (+)& 2.07e+01 $\pm$2.21e-02 (+)& 2.03e+01$\pm$6.80e-03 (+)& \textbf{6.58e+00$\pm$3.10e-01} (-)& 7.67e+00$\pm$3.10e-01 \\ \cline{2-10}
         & D2 & 2.11e+01$\pm$5.28e-01 (+)& 1.71e+01$\pm$5.22e-02 (+)& 2.12e+01$\pm$5.10e-01 (+)& 2.07e+01$\pm$4.33e-01 (+)& 2.03e+01$\pm$1.01e-01 (+)& 2.08e+01$\pm$1.60e-02 (+)& 1.24e+01$\pm$3.80e-01 (+)& \textbf{1.05e+01$\pm$3.10e-01} \\ \cline{2-10}
         & D3 & 2.11e+01$\pm$5.29e-01 (+)& 1.85e+01$\pm$4.45e-02 (+)& 2.12e+01$\pm$3.80e-01 (+)& 2.06e+01$\pm$4.33e-01 (+)& 2.01e+01 $\pm$2.38e-02 (+)& 2.07e+01$\pm$1.11e-02 (+)& 1.78e+01$\pm$5.40e-01 (+)& \textbf{1.71e+01$\pm$5.20e-01} \\ \cline{2-10}
         & D4 & 2.13e+01$\pm$5.32e-01 (+)& 1.86e+01$\pm$1.15e-02 (+)& 2.13e+01$\pm$4.20e-01 (+)& 2.09e+01$\pm$5.95e-01 (+)& 2.07e+01$\pm$ 2.81e-02 (+)& 2.09e+01$\pm$8.02e-03 (+)& 1.80e+01$\pm$4.60e-01 (+)& \textbf{1.75e+01$\pm$2.70e-01} \\ \cline{2-10}
         & D5 & 2.12e+01$\pm$5.31e-01 (+)& 1.85e+01$\pm$3.01e-02 (+)& 2.13e+01$\pm$4.90e-01 (+)& 2.08e+01$\pm$5.42e-01 (+)& 2.07e+01$\pm$ 4.91e-02 (+)& 2.09e+01$\pm$7.28e-03 (+)& 1.80e+01$\pm$3.70e-01 (+)& \textbf{1.73e+01$\pm$2.20e-01} \\
        \hline 

        \multirow{5}{*}{F7} 
         & D1 & 1.02e+00$\pm$2.55e-02 (+)& 5.28e-01$\pm$7.81e-03 (-)& 1.02e+00$\pm$1.50e-02 (+)& 1.00e+00$\pm$1.81e-02 (+)& 1.00e+00$\pm$9.01e-04 (+)& 1.01e+00$\pm$ 9.64e-04 (+)& \textbf{4.60e-01$\pm$1.90e-01} (-)& 5.90e-01$\pm$9.00e-02 \\ \cline{2-10}
         & D2 & 1.02e+00$\pm$2.56e-02 (+)& 8.90e-01 $\pm$3.38e-03 (+)& 1.02e+00$\pm$2.10e-02 (+)& 1.00e+00$\pm$1.74e-02 (+)& 1.01e+00 $\pm$1.88e-04 (+)& 1.01e+00$\pm$9.16e-04 (+)& \textbf{6.50e-01$\pm$7.00e-02} (-)& 7.50e-01$\pm$8.00e-02 \\ \cline{2-10}
         & D3 & 1.02e+00$\pm$2.56e-02 (+)& 9.99e-01$\pm$4.54e-04 (+)& 1.02e+00$\pm$1.80e-02 (+)& 9.96e-01$\pm$1.53e-02 (+)& 1.01e+00$\pm$1.18e-03 (+)& 1.01e+00$\pm$1.08e-03 (+)& 1.00e+00$\pm$1.00e-02 (+)& \textbf{9.20e-01$\pm$4.00e-02} \\ \cline{2-10}
         & D4 & 1.03e+00$\pm$2.57e-02 (+)& 9.47e-01$\pm$2.46e-03 ($\approx$)& 1.03e+00$\pm$1.20e-02 (+)& 1.00e+00$\pm$1.98e-02 (+)& 1.01e+00 $\pm$ 9.93e-04 (+)& 1.01e+00$\pm$1.33e-03 (+)& 9.60e-01$\pm$2.00e-02 (+)& \textbf{9.40e-01$\pm$2.00e-02} \\ \cline{2-10}
         & D5 & 1.03e+00$\pm$2.57e-02 (+)& \textbf{9.41e-01$\pm$9.05e-03} (-)& 1.03e+00$\pm$1.50e-02 (+)& 1.00e+00$\pm$1.38e-02 (+)& 1.01e+00 $\pm$1.59e-03 (+)& 1.02e+00$\pm$5.46e-04 (+)& 9.60e-01$\pm$2.00e-02 (-)& 9.80e-01$\pm$1.00e-02 \\
        \hline 

        \multirow{5}{*}{F8} 
         & D1 & 1.66e+02$\pm$4.14e+00 (+)& 1.03e+02$\pm$7.45e-01 (+)& 1.85e+02$\pm$5.45e+00 (+)& 1.79e+02$\pm$4.57e+00 (+)& 1.70e+02$\pm$9.17e-01 (+)& 1.56e+02$\pm$ 3.04e-01 (+)& \textbf{7.33e+01$\pm$7.72e+00} (-)& 8.93e+01$\pm$5.76e+00 \\ \cline{2-10}
         & D2 & 1.74e+02$\pm$4.35e+00 (+)& 1.25e+02$\pm$7.33e-01 (+)& 1.88e+02$\pm$6.12e+00 (+)& 1.79e+02$\pm$5.07e+00 (+)& 1.53e+02$\pm$1.18e+00 (+)& 1.69e+02 $\pm$2.76e-01 (+)& 1.16e+02$\pm$5.11e+00 (+)& \textbf{1.10e+02$\pm$1.70e+00} \\ \cline{2-10}
         & D3 & 1.72e+02$\pm$4.31e+00 (+)& 1.28e+02$\pm$3.83e-02 (+)& 1.86e+02$\pm$5.85e+00 (+)& 1.71e+02$\pm$4.49e+00 (+)& 1.57e+02 $\pm$6.61e-01 (+)& 1.65e+02$\pm$ 2.36e-01 (+)& 1.24e+02$\pm$3.43e+00 (+)& \textbf{1.17e+02$\pm$3.79e+00} \\ \cline{2-10}
         & D4 & 1.82e+02$\pm$4.55e+00 (+)& 1.43e+02$\pm$ 1.01e+00 (+)& 1.97e+02$\pm$7.10e+00 (+)& 1.70e+02$\pm$4.88e+00 (+)& 1.62e+02$\pm$1.63e+00 (+)& 1.75e+02$\pm$ 6.12e-01 (+)& 1.21e+02$\pm$7.42e+00 (+)& \textbf{1.15e+02$\pm$2.06e+00} \\ \cline{2-10}
         & D5 & 1.83e+02$\pm$4.58e+00 (+)& 1.42e+02$\pm$9.40e-01 (+)& 1.99e+02$\pm$6.90e+00 (+)& 1.74e+02$\pm$4.50e+00 (+)& 1.67e+02$\pm$1.68e+00 (+)& 1.75e+02$\pm$2.45e-01 (+)& 1.36e+02$\pm$6.03e+00 (+)& \textbf{1.22e+02$\pm$2.24e+00} \\
        \hline 
        \hline

        \multicolumn{2}{|c||}{\textbf{$+/\approx/-$}} & 39/1/0 & 34/4/2 & 39/1/0 & 38/2/0 & 37/2/1 & 39/1/0 & 31/3/6 & \textbf{NA} \\
        \hline
        \multicolumn{2}{|c||}{\textbf{Average Rank}} & 6.71 & 3.50 & 7.06 & 5.01 & 4.55 & 5.50 & 2.44 & \textbf{1.22} \\
        \hline
        \end{tabular}
        \vspace{-3em}
    }
\end{table*}
\subsubsection{Dynamic Tracking Behavior Analysis.} Figure \ref{fig:online_error_comparison} illustrates the online error trajectories. Following a drift, GeM-EA exhibits a ``cliff-like'' convergence profile, dropping almost vertically and stabilizing at a low error bound. This rapid recovery confirms the effectiveness of the bi-level meta-learned prior. Furthermore, thanks to the analytical solution for weights and linear residuals, GeM-EA maintains an $O(N)$ computational complexity, running significantly faster than GP-based methods like SAEF-1GP and comparable to the fastest baselines.

\subsection{Ablation \& Sensitivity \& Efficiency}
Due to strict space constraints, comprehensive ablation studies confirming the indispensability of each algorithmic component (BML, RAS, and GRE), detailed parameter sensitivity analyses on $P$ and $\tau$, and mathematical proofs of GeM-EA's linear computational complexity $O(N)$ alongside wall-clock runtime comparisons are presented in the Supplementary Appendix 2 and 3.

\section{Conclusion}
This paper presents GeM-EA, a meta-learning enhanced evolutionary algorithm for Streaming Data-Driven Optimization (SDDO) under concept drift. By synergizing bi-level surrogate adaptation and decoupling structural and non-structural parameters to alleviate gradient conflicts, together with a multi-island generative replay mechanism, GeM-EA achieves rapid optima tracking while effectively mitigating negative transfer. SDDObench experiments confirm its superior convergence, cliff-like recovery, and linear complexity. Future work will leverage automated design principles and unified benchmarking \cite{qiu2026automated,ma2025metaboxv2,guo2025designx} to scale GeM-EA toward higher-dimensional, many-objective environments with fully autonomous algorithm design capabilities.

\begin{acks}
This work was supported in part by the Guangdong Provincial Natural Science Foundation for Outstanding Youth Team Project (Grant No. 2024B1515040010), in part by Guangzhou Science and Technology Elite Talent Leading Program for Basic and Applied Basic Research (Grant No. SL2024A04J01361), in part by the Fundamental Research Funds for the Central Universities (Grant No. 2025ZYGXZR027).
\end{acks}
\bibliographystyle{ACM-Reference-Format}
\bibliography{ref}

\end{document}